\documentclass[letterpaper, 10 pt, conference]{ieeeconf}

\IEEEoverridecommandlockouts                              % This command is only needed if 
\usepackage{graphicx} % for pdf, bitmapped graphics files
\usepackage{epsfig} % for postscript graphics files
\usepackage{epstopdf}
\usepackage{dblfloatfix} % fix for bottom-placement of figure
\usepackage{courier}

\usepackage{color}
\usepackage{amsmath} % cmex10
\usepackage{amssymb}
\usepackage{float}
\usepackage[backend=bibtex,bibstyle=ieee,citestyle=numeric-comp]{biblatex}
\newtheorem{theorem}{Theorem}[section]
\newtheorem{definition}[theorem]{Definition}

\usepackage{balance}
\title{\LARGE \bf
Invariant Extended Kalman Filtering Using Two Position Receivers for Extended Pose Estimation
}
\author{Natalia Pavlasek$^{1}$, Alex Walsh$^{2}$, and James Richard Forbes$^{3}$% <-this % stops a space
\thanks{*This work was supported by the William Dawson Scholar, NSERC Discovery Grant, and Candadian Foundation for Innovation JELF programs.
}% <-this % stops a space
\thanks{$^{1}$Department of Mechanical Engineering, McGill University, 817 Sherbrooke St. W., Montreal, QC,
Canada, H3A 0C3. e-mail: {\tt\small natalia.pavlasek@mail.mcgill.ca}}%
\thanks{$^{2}$No affiliation. e-mail: 
        {\tt\small alex.walsh@mail.mcgill.ca}}%
\thanks{$^{3}$Department of Mechanical Engineering, McGill University, 817 Sherbrooke St. W., Montreal, QC,
Canada, H3A 0C3. e-mail: 
        {\tt\small james.richard.forbes@mcgill.ca}}%
\vspace{-1cm}}

\newcommand{\ignore}[1]{}

\newcommand{\mc}[1]{\mathcal{#1}}

\newcommand{\bma}[1]{\left[\begin{array}{ #1}}
\newcommand{\ema}{\end{array}\right]}

\DeclareMathAlphabet{\mbf}{OT1}{ptm}{b}{n}
\newcommand{\mbs}[1]{{\boldsymbol{#1}}}
\newcommand{\mbsdot}[1]{{\dot{\boldsymbol{#1}}}}

\newcommand{\mbfdot}[1]{{\dot{\mbf{#1}}}}
\newcommand{\mbfbar}[1]{{\bar{\mbf{#1}}}}
\newcommand{\mbfhat}[1]{{\hat{\mbf{#1}}}}

\def\fdotb{{\raisebox{-0.6ex}{ \kern0.2ex\raisebox{0.8ex}{\tiny $\hspace*{-1ex}\circ$}}}}
\def\fddotb{{\raisebox{-0.6ex}{ \kern0.2ex\raisebox{0.8ex}{\tiny $\hspace*{-1ex}\circ\circ$}}}}

\newcommand{\trans}{{\ensuremath{\mathsf{T}}}} % transpose
\newcommand{\utimes}{ {\raisebox{-0.6ex}{ \kern-1.0ex\raisebox{0.6ex}{ \small $\mathsf{v}$}}} } % 
\newcommand{\beq}{\begin{equation}}
\newcommand{\eeq}{\end{equation}}
\newcommand{\bdis}{\begin{displaymath}}
\newcommand{\edis}{\end{displaymath}}
\newcommand{\beqarray}{\begin{eqnarray}}
\newcommand{\eeqarray}{\end{eqnarray}}
\newcommand{\beqarraynn}{\begin{eqnarray*}}
\newcommand{\eeqarraynn}{\end{eqnarray*}}
\newcommand{\balign}{\begin{align}}
\newcommand{\ealign}{\end{align}}
\newcommand{\balignnn}{\begin{align*}}
\newcommand{\ealignnn}{\end{align}}

\renewcommand{\theenumii}{\arabic{enumii}}
\makeatletter
\renewcommand{\p@enumii}{\theenumi.}
\makeatother
\begin{document}
%auto-ignore
% This is not a standalone latex document. To use this file
% as a cover page on an arXiv upload of a document that is 
% already accepted as some sort of IEEE publication, you must
%
%  1) add the following just after the \begin{document} line
%     of your main paper document
%
%         \input{arxiv-cover-ieee.tex}
%
%  2) and replace the relevant information in the block below.
%
% The relevant information has been parameterized as variables.
% Simply replace the variable values with your stuff and the 
% result should be good.
%
% Make sure to not include this file for ACTUAL submissions to 
% the IEEE. Luckily you can just comment in/out the 
% \input{arxiv-cover-ieee.tex} line.
%
% FYI: The exact citation with formatting can be obtained 
% from your paper's page on IEEE Xplore.
%
%%%%%%%%%%%%%%%%%%%%%%%%%%%%%%%%%%%%%%%%%%%%%%%%%%%%%%%%%%%%%%%
%%%%%%%%%%%%%%%%%%%%%% ADD YOUR INFO HERE %%%%%%%%%%%%%%%%%%%%%
%%%%%%%%%%%%%%%%%%%%%%%%%%%%%%%%%%%%%%%%%%%%%%%%%%%%%%%%%%%%%%%
\def \myJournal {IEEE International Conference on Robotics and Automation 2021}
\def \myDoi {}
\def \myPaperSiteName {IEEE Xplore}
\def \myPaperSiteLink {}
\def \myYear {2021}
\def \myPaperCitation{N. Pavlasek, A. Walsh, and J. R. Forbes, ``Invariant Extended Kalman Filtering Using Two Position Receivers for Extended Pose Estimation,'' in \textit{IEEE International Conference on Robotics and Automation}, 2021.}

%%%%%%%%%%%%%%%%%%%%%%%%%%%%%%%%%%%%%%%%%%%%%%%%%%%%%%%%%%%%%%%
%%%%%%%%%%%%%%%%%%%%%%%%%%%%%%%%%%%%%%%%%%%%%%%%%%%%%%%%%%%%%%%

\begin{figure*}[t]

\thispagestyle{empty}
\begin{center}
\begin{minipage}{6in}
\centering
This paper has been accepted for presentation at \emph{\myJournal}. 
\vspace{1em}

This is the author's version of an article that has, or will be, published in this journal or conference. Changes were, or will be, made to this version by the publisher prior to publication.
\vspace{2em}

%\begin{tabular}{rl}
%DOI: & \myDoi\\
%\myPaperSiteName: & \texttt{\myPaperSiteLink}
%\end{tabular}

%\vspace{2em}
Please cite this paper as:

\myPaperCitation

\vspace{15cm}
\copyright \myYear \hspace{4pt}IEEE. Personal use of this material is permitted. Permission from IEEE must be obtained for all other uses, in any current or future media, including reprinting/republishing this material for advertising or promotional purposes, creating new collective works, for resale or redistribution to servers or lists, or reuse of any copyrighted component of this work in other works.

\end{minipage}
\end{center}
\end{figure*}
\newpage
\clearpage
\pagenumbering{arabic}

\maketitle
\thispagestyle{empty}
\pagestyle{empty}

%%%%%%%%%%%%%%%%%%%%%%%%%%%%%%%%%%%%%%%%%%%%%%%%%%%%%%%%%%%%%%%%%%%%%%%%%%%%%%%%
\begin{abstract}
% This work presents a solution to the state estimation problem that uses two position receivers and an inertial measurement unit (IMU) to produce position and attitude corrections. 

% The novelty of the proposed navigation framework lies in how the two position receivers are used.

% using an invariant extended Kalman filter (IEKF).

This paper considers the use of two position receivers and an inertial measurement unit (IMU) to estimate the position, velocity, and attitude of a rigid body, collectively called extended pose. The measurement model consisting of the position of one receiver and the relative position between the two receivers is left invariant, enabling the use of the invariant extended Kalman filter (IEKF) framework.
% The specific form of the navigation problem proposed is shown to be compatible with the invariant extended Kalman filter (IEKF) framework. In particular, the measurement model consists of a position receiver along with the relative position of the two position receivers. It is shown that the proposed measurement model is left invariant, enabling the use of the IEKF framework. 
The IEKF possesses various advantages over the standard multiplicative extended Kalman filter, such as state-estimate-independent Jacobians. Monte Carlo simulations demonstrate that the two-receiver IEKF approach yields improved estimates over a two-receiver multiplicative extended Kalman filter (MEKF) and a single-receiver IEKF approach. An experiment further validates the proposed approach, confirming that the two-receiver IEKF has improved performance over the other filters considered.

% , and along with the satisfaction of additional technical criteria, enables application of the IEKF framework. 

% The IEKF possesses various advantages over the standard multiplicative extended Kalman filter, such as state-estimate-independent Jacobians.

% The filter state is defined as an element of $SE_2(3)$, the group of double direct isometries. The prediction step 

% The IMU is used in the IEKF prediction step to propagate the state estimate, which is 

% The prediction step is posed on matrix Lie group $SE_2(3)$. The correction step uses 

% The measurement model consists of a position receiver and with the relative position of the two position receivers. The measurement model is shown to be left invariant. 

% The state estimation problem is posed on the on the matrix Lie group $SE_2(3)$. The prediction step 

% Additionally, using one position receiver, and the relative position of the two receivers, 

% The proposed approach is presented on the matrix Lie group $SE_2(3)$ using an invariant extended Kalman filter (IEKF). The IEKF has advantages over the standard multiplicative extended Kalman filter such as state independent Jacobians, provided certain conditions are met. Monte Carlo simulations demonstrate that the two-receiver IEKF approach yields improved estimates over a two-receiver MEKF and a single-receiver IEKF approach. An experiment further validates the proposed approach, confirming that the two-receiver IEKF has improved performance over the other filters.
\end{abstract}

%%%%%%%%%%%%%%%%%%%%%%%%%%%%%%%%%%%%%%%%%%%%%%%%%%%%%%%%%%%%%%%%%%%%%%%%%%%%%%%%
\section{Introduction}

Estimating robot position, velocity, and attitude, collectively referred to as extended pose \cite{Barrau_Bonnabel_ICRA_2020}, is typically accomplished by fusing interoceptive and exteroceptive sensor data together using a nonlinear filter. Common interoceptive sensors are accelerometers and rate gyros. GPS receivers, or more generally GNSS receivers, ultra-wideband (UWB) receivers, and long baseline (LBL) acoustic arrays are common exteroceptive positioning sensors used for outdoor, indoor, and underwater navigation, respectively. The problem of estimating extended pose using an accelerometer, a rate gyro, and two position receivers, referred to as the two-receiver extended pose estimation problem, is the focus of this paper. This sensor suite is widely available, has improved observably properties relative to a one-receiver set-up, and depending on the application domain, the navigation solution proposed herein can be applied in outdoor, indoor, and underwater scenarios. 

The two-receiver extended pose estimation problem has many solutions that can be split into tightly-coupled and loosely-coupled approaches. Tightly coupled approaches such as the carrier-phase approach \cite{Cohen1996,Hirokawa2009,Buist2009,Emelyantsev2018} or pseudorange approach~\cite{Shin2005} are effective, but add complexity. For example, resolution of the integer ambiguity problem is required~\cite{Buist2009}. Loosely-coupled receiver-INS solutions are simpler to implement, and are effective for rapid deployment of estimation algorithms.
% A loosely-coupled GPS-INS solution for two receivers is to use GPS velocity measurements with two receivers and an IMU in an extended Kalman filter (EKF)~\cite{Yoon2013}. The GPS-velocity approach determines angles using trigonometry manually, which is cumbersome compared to~\cite{Hao2018}, where position measurements between two-antenna are used directly. 
While solutions using extended Kalman filters (EKFs) exist for the loosely-coupled problem~\cite{Yoon2013,Hao2018}, new developments in state estimation, such as the invariant EKF (IEKF)~\cite{Barrau2017TAC}, warrant a second look at the two-receiver problem.

% The Bayes filter \cite{Sarkka_2013} is often approximated in practice resulting in the particle filter, a sigma point Kalman filter such as the unscented Kalman filter (UKF), or the EKF. 

% sigma point Kalman filter (SPKFs)
The particle filter (PF), the unscented Kalman filter (UKF), the EKF, and many other filters, are approximations of the Bayes filter \cite{Sarkka_2013}. The IEKF is a variant of the EKF that takes advantage of the symmetry of matrix Lie groups that naturally arises in attitude, pose (i.e., position and attitude), and extended pose estimation problems. Several desirable properties arise from the symmetry-preserving nature of the system. For instance, provided certain conditions are met, the IEKF state-estimation error is independent of the true vehicle trajectory~\cite{Barrau2017}. Moreover, the IEKF is guaranteed to converge locally under certain assumptions~\cite{Barrau2017}. The IEKF has improved transient response over the MEKF, especially with large initial errors~\cite{Barrau2017,Barrau2017TAC,Arsenault2019}. As such, the goal of this paper is to leverage the IEKF in the two-receiver extended pose estimation problem.
The novel contribution of this paper is demonstrating how to use an accelerometer, a rate gyro, and two position receivers to estimate position, velocity, and attitude within the IEKF framework. The proposed two-receiver IEKF is compared to a two-receiver MEKF, similar to \cite{Shin2005,Hao2018}, in both simulation and experiment. The experiments presented use position measurements from a UWB system. UWB is a highly accurate positioning system that can be used in GPS denied environments~\cite{Richardson2010}, and that does not suffer from long-term drift~\cite{Nageli2016}.

% In addition, this work compares the two-receiver problem in the IEKF framework to the corresponding MEKF approach, presented in~\cite{Shin2005} and~\cite{Hao2018}, in both simulation and experimentation.

% The novel concept of this paper is the resolution of the two position receiver problem in the invariant framework. This is shown to enhance state estimates over the corresponding MEKF approach, when the initialization error is large, and over the equivalent single-receiver problem. The result is first verified using Monte Carlo simulations. An experiment is then used to demonstrate the effectiveness of the proposed approach on real data.

The remainder of this paper is organized as follows. Section \ref{sec:prelim} reviews some preliminaries, namely matrix Lie group theory, the IEKF equations, and the average normalized innovation squared (NIS) test. Section \ref{sec:2receiver} introduces the problem and summarizes the equations of motion before framing the problem in the IEKF framework. Section \ref{sec:results} presents the findings from simulations and an experiment. Finally, Section \ref{sec:conclusion} provides concluding remarks.

\section{Preliminaries} \label{sec:prelim}

\subsection{Matrix Lie Groups}
Consider  the  matrix  Lie  group $\mathcal{G}$ composed of $n\times n$ invertible matrices with $m$ degrees of freedom that is closed under matrix multiplication \cite{Hall2014}. The matrix Lie algebra associated with $\mathcal{G}$ is denoted $\mathfrak{g}$. An element of $\mathfrak{g}$ can be mapped to $\mathcal{G}$ using the exponential map, $\exp(\cdot) : \mathfrak{g} \rightarrow \mathcal{G}$. Similarly, an element of $\mathcal{G}$ can be mapped to $\mathfrak{g}$ using the matrix natural logarithm, $\log(\cdot) : \mathcal{G} \rightarrow \mathfrak{g}$. The linear operator $(\cdot)^\wedge$ maps $\mathbb{R}^m \to \mathfrak{g}$, and $(\cdot)^\vee$ maps $\mathfrak{g} \to \mathbb{R}^m$. When $\delta \mbs{\xi} \in \mathbb{R}^m$ is small, $\mathrm{exp} \left(\delta \mbs{\xi}^\wedge \right)$ can be approximated as $\mathrm{exp} \left(\delta \mbs{\xi}^\wedge \right) \approx \mbf{1} + \delta \mbs{\xi}^\wedge$.

%, where $\dim \mathfrak{g}$ is the dimension of $\mathfrak{g}$.

% The matrix Lie algebra associated with $\mathcal{G}$, denoted $\mathfrak{g}$, is defined as $T_\mbf{1}\mathcal{G}$, and represents the tangent space of $\mathcal{G}$ at the identity, denoted $\mbf{1}$. Elements of the matrix Lie algebra can be mapped to the matrix Lie group using the exponential map, $\exp(\cdot) : \mathfrak{g} \rightarrow \mathcal{G}$. The inverse mapping uses the matrix natural logarithm, $\log(\cdot) : \mathcal{G} \rightarrow \mathfrak{g}$. Elements of the matrix Lie algebra can be mapped to a $m$ dimensional column matrix using the linear operator $(\cdot)^\vee : \mathfrak{g} \rightarrow \mathbb{R}^m$. The inverse mapping is $(\cdot)^\wedge : \mathbb{R}^m \rightarrow \mathfrak{g}$.

\subsection{Invariant Extended Kalman Filtering Preliminaries}
\label{sec:iekf_prelim}

The IEKF framework has four conditions that must be met for it to be used. First, the measurement model must be either left or right invariant. The second and third conditions are that the invariant error and invariant innovation must be consistent with the left or right invariance of the measurement model. Fourth, the process model must be group affine. In this section, each of these components will be discussed, with a focus on left invariance.

% Second, the invariant error, and third, the innovation form, must be consistent with the left- or right-invariance of the measurement model. Fourth, the process model must be group affine. In this section, each of these components will be discussed, with a focus on left invariance.

\begin{definition}[Left-Invariant Measurement Model~\cite{Barrau2017TAC}]
\label{def:left_invar_meas}
Consider a measurement $\mbf{y}_k \in \mathbb{R}^{n}$, a state $\mbf{X}_k \in \mathcal{G} \subset \mathbb{R}^{n \times n} $, where $\mathcal{G}$ is a matrix Lie group, a known column matrix $\mbf{b} \in \mathbb{R}^n$, and zero-mean Gaussian white noise $\mbf{n}_k \in \mathbb{R}^{n}$. A left-invariant measurement model is defined by
\begin{align} \label{eq:left_inv_meas}
	\mbf{y}_k &= \mbf{X}_k \mbf{b} + \mbf{n}_k.
\end{align}
\end{definition}

\begin{definition}[Left-Invariant Error~\cite{Barrau2017TAC}]
\label{def:left_invar_error}
Let $\mbf{X}_k \in \mathcal{G}$ be the true state of a system, $\mbfbar{X}_k \in \mathcal{G}$ be a state different from the true state, such as the predicted or corrected state estimates. The left invariant error is defined by
\begin{align} \label{eq:left_error}
    \delta \mbf{X}_k = \mbf{X}_k^{-1} \mbfbar{X}_k.
\end{align}
\end{definition}

\begin{definition}[Left-Invariant Innovation~\cite{Barrau2017TAC}]
\label{def:left_invar_innovation}
Let $\check{\mbf{y}}_k \in \mathbb{R}^{n}$ be the predicted measurement, and $\mbf{z}_k \in \mathbb{R}^{n}$ be the innovation. The left-invariant innovation is given by
\begin{align} \label{eq:left_inn}
    \mbf{z}_k = \check{\mbf{X}}_k^{-1}(\mbf{y}_k - \check{\mbf{y}}_k).
\end{align}
\end{definition}

% :\mathcal{G} \times \mathbb{R}^{n_u} \to \mathcal{G}
\begin{definition}[Group Affine~\cite{Barrau2017TAC}]
\label{def:group_affine}
A function $\mbf{F}(\mbf{X},\mbf{u})$ is said to be group affine if for all $\mbf{X}_1, \mbf{X}_2 \in \mathcal{G}$ and $ \mbf{u} \in \mathbb{R}^{n_u}$, the equation
\begingroup\makeatletter\def\f@size{10}\check@mathfonts
\def\maketag@@@#1{\hbox{\m@th\large\normalfont#1}}%
\begin{align} 
    \hspace{-10pt}\mbf{F}(\mbf{X}_1\mbf{X}_2,\mbf{u}) = \mbf{X}_1\mbf{F}(\mbf{X}_2,\mbf{u}) + \mbf{F}(\mbf{X}_1,\mbf{u})\mbf{X}_2 - \mbf{X}_1\mbf{F}(\mbf{1},\mbf{u})\mbf{X}_2
\end{align}\endgroup
is satisfied.
\end{definition}

Given the left-invariant measurement model Definition~\ref{def:left_invar_meas}, error given by Definition~\ref{def:left_invar_error}, innovation from Definition~\ref{def:left_invar_innovation}, and a process model that satisfies Definition~\ref{def:group_affine}, the IEKF framework can be used~\cite{Barrau2017TAC}.

%
% The final requirement for the IEKF is a group affine process model. A function $\mbf{F}(\mbf{X},\mbf{u})$ is said to be group affine if \cite{Barrau2017TAC}
% %
% \begin{align}
%     \mbf{F}(\mbf{X}_1\mbf{X}_2,\mbf{u}) = \mbf{X}_1\mbf{F}(\mbf{X}_2,\mbf{u}) + \mbf{F}(\mbf{X}_1,\mbf{u})\mbf{X}_2 - \mbf{X}_1\mbf{F}(\mbf{1},\mbf{u})\mbf{X}_2
% \end{align}
% %
% is satisfied, where $\forall \mbf{X}_1, \mbf{X}_2 \in \mathcal{G}$ and $\forall \mbf{u} \in \mathbb{R}^{n_u}$ 

% A multiplicative uncertainty can be defined as
% %
% \begin{align}
%     \mbf{X} = \mbfbar{X} \exp(\delta \mbs{\xi}^\wedge)  ,
% \end{align}
% %
% where $\delta \mbs{\xi} \in \mathbb{R}^m$. This is, however, inconsistent with the left-invariant error definition. This can be remedied by negating the measurement Jacobian, which in turn negates the Kalman gain. 

% The left-invariant correction is then,
% %
% \begin{align}
%     \mbf{X} = \mbfbar{X} \exp(-\delta \mbs{\xi}^\wedge)  ,
% \end{align}
% %

% \emph{Definition 1 (Group Affine \cite{Barrau2017TAC}):} 

% Provided the measurement is left invariant, the error is left invariant, the innovation is left invariant, and the process model is group affine, the IEKF framework can be employed. The IEKF prediction and correction steps can be defined as follows.

\subsection{Invariant Extended Kalman Filtering Implementation}

% Given the IEKF preliminaries from Section~\ref{sec:iekf_prelim}, the IEKF prediction and correction equations can be presented \cite{Barrau2017TAC}. 
Consider a process model of the form
\begin{align} 
	\mbfdot{X}(t) &= \mbf{F}(\mbf{X}(t),\mbf{u}(t)) + \mbf{X}(t) \mbf{W}(t), \label{eq:nonlin_CT_proc_model}
\end{align}
where $\mbf{F}(\mbf{X}(t),\mbf{u}(t))$ is group affine, $\mbf{X}(t) \in \mathcal{G}$, $\mbf{u}(t) \in \mathbb{R}^{n_u}$ is the process input, and $\mbf{W}(t) = \mbf{w}(t)^\wedge \in \mathfrak{g}$ is zero-mean Gaussian white process noise. For the remainder of this work, the argument~$(t)$ is dropped for conciseness, unless required for clarity. The nonlinear process model can be linearized using the definition of the left-invariant error \eqref{eq:left_error} to yield
\begin{align}
    \delta \mbsdot{\xi} &= \mbf{A}_\mathrm{c} \delta \mbs{\xi} + \mbf{L}_\mathrm{c} \delta \mbf{w} , \label{eq:proc_model} 
\end{align}
% Both the \eqref{eq:nonlin_CT_proc_model} and \eqref{eq:proc_model} can be distritized
where $\delta \mbf{w} \sim \mathcal{N}(\mbf{0}, \mbs{Q} \delta(t - \tau))$, and $\mbs{Q}$ is the power spectral density.
The discrete time prediction is then
\begin{align}
    \check{\mbf{X}}_k &= \mbf{F}_{k-1}\big( \mbfhat{X}_{k-1}, \mbf{u}_{k-1} \big), \label{eq:nonlin_DT_proc_model_pred} \\
	\check{\mbf{P}}_k &= \mbf{A}_{k-1} \mbfhat{P}_{k-1}\mbf{A}_{k-1}^\trans + \mbf{L}_{k-1}\mbf{Q}_{k-1}\mbf{L}_{k-1}^\trans,
\end{align}
where \eqref{eq:nonlin_DT_proc_model_pred} comes from discretization of \eqref{eq:nonlin_CT_proc_model}, and $\mbf{A}_{k-1}$, $\mbf{L}_{k-1}$, and $\mbf{Q}_{k-1}$ come from discretization of \eqref{eq:proc_model} and $\mbs{Q}$  \cite{VanLoan1978}.
The left-invariant innovation can be linearized, again using the definition of the left-invariant error \eqref{eq:left_error}, along with the approximation $\delta \mbf{X}_k = \exp(\delta \mbs{\xi}_k^\wedge ) \approx \mbf{1} + \delta \mbs{\xi}_k^\wedge$. % , resulting in
In detail,
\begin{align}
    \mbf{z}_k % &= \check{\mbf{X}}_k^{-1}(\mbf{y}_k - \check{\mbf{y}}_k) \nonumber\\
    &= \check{\mbf{X}}_k^{-1}(\mbf{X}_k\mbf{b} + \mbf{n}_k - \check{\mbf{X}}_k \mbf{b}) 
    = \delta \mbf{X}_k^{-1}\mbf{b} - \mbf{b} + \check{\mbf{X}}_k^{-1} \mbf{n}_k \nonumber\\
    &\approx (\mbf{1}-\delta \mbs{\xi}_k^\wedge )\mbf{b} - \mbf{b} + \check{\mbf{X}}_k^{-1} \delta \mbf{n}_k 
     % &= -\delta \mbs{\xi}_k^\wedge \mbf{b} + \check{\mbf{X}}_k^{-1}\mbf{n}_k \nonunmber \\
      = \mbf{H}_k \delta \mbs{\xi}_k + \mbf{M}_k \delta \mbf{n}_k . \label{eq:lin_inn}
\end{align}
%
% \begin{align}
%   \mbf{z}_k &= \mbf{H}_k \delta \check{\mbs{\xi}}_k + \mbf{M}_k \delta \mbf{n}_k, \label{eq:lin_inn}
% \end{align}
%
where $\mbf{M}_k = \check{\mbf{X}}_k^{-1}$ and $\delta \mbf{n}_k \sim \mathcal{N}\left(\mbf{0},\mbf{R}_{k}\right)$. %  and $\mbf{R}_{k}$ is the measurement noise covariance.
The state correction is then \vspace{-5pt}
\begin{align}
	\mbfhat{X}_k = \check{\mbf{X}}_k \exp \Big( -(\mbf{K}_k\mbf{z}_k)^\wedge \Big) ,
\end{align}
which is consistent with the definition of the left-invariant error in \eqref{eq:left_error}. The Kalman gain is
\begin{align}
    \mbf{K}_k = \check{\mbf{P}}_k\mbf{H}_k^\trans \mbf{S}_k^{-1} , \label{eq:Gain_forward} 
\end{align}
where 
\begin{align}
    \mbf{S}_k
    =
    \mbf{H}_k\check{\mbf{P}}_k\mbf{H}_k^\trans+\mbf{M}_k\mbf{R}_k\mbf{M}_k^\trans  . \label{eq:innov_gain} 
\end{align}
% The left-invariant error is defined as
% %
% \begin{align}
%     \delta \mbf{X}= \mbf{X}^{-1}\mbfhat{X}.
% \end{align}

\subsection{Normalized Innovation Squared Test}
% also termed Mahalanobis distance

% (\color{red}{check and reference}\color{black})
The normalized innovation squared (NIS) is defined as \cite{BarShalom2001,Zhaozhong2018}
\begin{align}
	\epsilon_{z,k} = \mbf{z}_k^\trans \mbf{S}_k^{-1} \mbf{z}_k, 
\end{align}
where $\mbf{S}_k$ is given in \eqref{eq:innov_gain}.
The NIS is $\chi^2$ distributed of degree $n_{z}$. The average NIS, 
\begin{align}\label{eq:ANIS}
	\bar{\epsilon}_{z,k} = \frac{1}{N} \sum_{i=1}^{N} \epsilon^i_{z,k},
\end{align}
is also $\chi^2$ distributed of degree $Nn_{z}$, and can therefore be compared to the critical $\chi^2$ values to verify the consistency of the estimator \cite{BarShalom2001}. Dividing the result of \eqref{eq:ANIS} by $n_{z}$ results in a values close to one for consistent systems \cite{Barrau2015b}. A value greater than one indicates an optimistic estimator, where the errors to covariance ratio is too large, or the estimates include significant bias \cite{BarShalom2001}. A value less than one indicates that the estimator is pessimistic.

\section{The Two Receiver Problem} \label{sec:2receiver}

\iffalse
Goal: Estimate both position and attitude using two position measurements JRF - and an accel and rate gyro. 
\fi

\subsection{Notation}
The following notation is used throughout this work. A physical vector $\underrightarrow{p}$ resolved in $\mathcal{F}_a$ is denoted $\mbf{p}_a$, and resolved in $\mathcal{F}_b$ is denoted as $\mbf{p}_b$. The relationship between $\mbf{p}_a$ and $\mbf{p}_b$ is $\mbf{p}_a = \mbf{C}_{ab}\mbf{p}_b$, or $\mbf{p}_b = \mbf{C}_{ba}\mbf{p}_a$, where $\mbf{C}_{ab} = \mbf{C}_{ba}^\trans = \mbf{C}_{ba}^{-1} \in SO(3)$ is the direction cosine matrix (DCM). The cross product of two physical vectors resolved in a frame, such as $\mc{F}_b$, is denoted $\mbf{p}_b^\times \mbf{q}_b = -\mbf{q}_b^\times \mbf{p}_b$. 

\subsection{Problem Formulation}
% , and the inverse of $\mbf{C}_{ab}$ is $\mbf{C}_{ba} = \mbf{C}_{ab}^\trans$.

% \cite{Bernstein2008}
% in $\mathcal{F}_a$
Consider a rigid body, with body frame $\mc{F}_b$, an inertial frame $\mc{F}_a$, and a datum point $w$. An IMU with an accelerometer and a rate gyro is mounted at point $z_0$ on the rigid body. Two position receivers, which could be GPS or UWB receivers, are placed at points $z_1$ and $z_2$, respectively, on the rigid body. The physical vector describing $z_0$ relative to $w$ is $\underrightarrow{r}^{z_0 w}$, and the physical vector describing $z_1$ relative to $z_2$ is denoted $\underrightarrow{r}^{z_1z_2}$. The velocity of point $z_0$ relative to point $w$ relative to $\mc{F}_a$ is $\underrightarrow{v}^{z_0 w / a}$.

% Consider a reference frame $\mathcal{F}_a$ and a body-fixed frame $\mathcal{F}_b$. 
% The direction cosine matrix (DCM) relating $\mathcal{F}_a$ and $\mathcal{F}_b$ is given by $\mbf{C}_{ab} \in SO(3)$. DCMs are orthonormal, meaning $\mbf{C}_{ab}^\trans \mbf{C}_{ab} = \mbf{1}$, and as a result, $\mbf{C}_{ab}^\trans = \mbf{C}_{ab}^{-1}$. 

%A physical vector $\underrightarrow{u}$ resolved in $\mathcal{F}_a$ is denoted $\mbf{u}_a$, and resolved in $\mathcal{F}_b$ is denoted as $\mbf{u}_b$. The relationship between $\mbf{u}_a$ and $\mbf{u}_b$ is $\mbf{u}_a = \mbf{C}_{ab}\mbf{u}_b$, or $\mbf{u}_b = \mbf{C}_{ba}\mbf{u}_a$, where $\mbf{C}_{ab} = \mbf{C}_{ba}^\trans = \mbf{C}_{ba}^{-1} \in SO(3)$ is the direction cosine matrix (DCM). The cross product of two physical vectors resolved in a frame, such as $\mc{F}_b$, is denoted $\mbf{u}_b^\times \mbf{v}_b = -\mbf{v}_b^\times \mbf{u}_b$. 

The two-receiver extended pose estimation problem to be solved is, given an accelerometer, a rate gyro, and two position receivers, estimate the position, velocity, and attitude as described by $\mbf{r}_a^{z_0 w}$, $\mbf{v}_a^{z_0 w / a}$, and $\mbf{C}_{ab}$, respectively. This problem follows a loosely-coupled GPS-INS formulation~\cite{BarShalom2001}.

\subsection{Measurement Models}
% Consider an IMU with a rate gyro and an accelerometer corrupted by zero-mean Gaussian white noise of the form $\mbf{w}_b^1 \sim \mathcal{N}(0,\mbs{Q}^\omega(t) \mathcal{\delta}(t-\tau))$ and $\mbf{w}_b^2 \sim \mathcal{N}(0,\mbs{Q}^\mathrm{a}(t) \mathcal{\delta}(t-\tau))$, respectively. 

The rate gyro, $\mbf{u}^1_{b}$, and accelerometer, $\mbf{u}_b^2$, measurement models are
% The interoceptive measurements are of the form
%
\begin{align}
	\mbf{u}^1_{b} &= \mbs{\omega}_b^{ba} - \mbf{w}^1_{b}, \label{eq:rate_gyro}\\
	\mbf{u}_b^2 &= \mbf{C}_{ab}^\trans ( \mbfdot{v}_a^{z_0w/a} - \mbf{g}_a ) - \mbf{w}_b^2,  \label{eq:accel}
	% \mbf{u}_b^2 &= \mbf{a}_b^{z_0w} - \mbf{w}_b^2,
\end{align}
where $\mbs{\omega}_b^{ba}$ is the angular velocity of $\mc{F}_b$ relative to $\mc{F}_a$ resolved in $\mc{F}_b$, $\mbfdot{v}_a^{z_0w/a}$ is the acceleration of point $z_0$ relative to point $w$ with respect to $\mc{F}_a$ resolved in $\mc{F}_a$, $\mbf{g}_a$ is the gravity vector resolved in $\mc{F}_a$, and both $\mbf{w}_b^1 \sim \mathcal{N}(0,\mbs{Q}^\omega(t) \mathcal{\delta}(t-\tau))$ and $\mbf{w}_b^2 \sim \mathcal{N}(0,\mbs{Q}^\mathrm{a}(t) \mathcal{\delta}(t-\tau))$ are zero-mean Gaussian white measurement noise.
Using \eqref{eq:rate_gyro} and \eqref{eq:accel} the continuous-time kinematic equations defining the time evolution of the rigid body are
\begin{align}
	\mbfdot{C}_{ab} &= \mbf{C}_{ab} (\mbf{u}_b^1 + \mbf{w}_b^1)^\times,\label{eq:kin_C} \\
	\mbfdot{v}_a^{z_0w/a} &= \mbf{C}_{ab}(\mbf{u}_b^2 + \mbf{w}_b^2) + \mbf{g}_a, \label{eq:kin_v}\\
	\mbfdot{r}^{z_0w}_a &= \mbf{v}_a^{z_0w/a}. \label{eq:kin_r}
\end{align}
Biases are not included in \eqref{eq:rate_gyro} and \eqref{eq:accel}, nor \eqref{eq:kin_C}, \eqref{eq:kin_v}, \eqref{eq:kin_r}, because it is assumed that the sensor data has been calibrated such that biases are negligible, which is the case in the experiments presented in Section~\ref{sec:experiments}. Biases can be included in the ``imperfect" IEKF framework \cite{Barrau2015,Heo:2018aa}.

% Section~\ref{sec:experiments
% because the measurement data used in the experiments presented in Section~\ref{sec:experiments} was calibrated and biases were deemed to be negligible after verification. Biases can be included necessitating the use of an ``imperfect" IEKF \cite{Barrau2015,Heo:2018aa}.

% Bias in the  rate gyro and accelerometer measurements are omitted because they were found to be negligible over the span considered in the experiments presented in Section~\ref{sec:experiments}. Bias states can be added to 

% The consideration of bias destroys the group affine properties of the process model, however the ``Imperfect IEKF'' can still be used, which retains some of the attractive properties of the IEKF \cite{Barrau2015}.

% These equations are discretized using a forward-Euler scheme.

% exteroceptive measurements
Position measurements from two position receivers affixed to the rigid body are available. Each position measurement is of the form 
\begin{align}
    \mbf{y}_k^{\textrm{pos},i} &= \mbf{r}^{z_iw}_{a_k} + \mbf{n}_k^{\textrm{pos},i}, \; \; \; i = 1,2, \label{eq:pos_meas}
\end{align}
where $\mbf{n}_k^i \sim \mathcal{N}(\mbf{0},\mbf{R}^{z_iw})$. 
Subtracting the two position measurements yields a relative position measurement, 
\begin{align}
        \mbf{y}_k^\textrm{rel} &= \mbf{y}_k^{\textrm{pos},2} - \mbf{y}_k^{\textrm{pos},1} = \mbf{r}^{z_2w}_{a_k} - \mbf{r}^{z_1w}_{a_k} + \mbf{n}_k^\textrm{rel} , \label{eq:rel_pos_meas}
\end{align}
where $\mbf{n}_k^\textrm{rel} = \mbf{n}_k^{\textrm{pos},2} - \mbf{n}_k^{\textrm{pos},1}$.
Notice that $\mbf{y}_k^\textrm{rel}$ in \eqref{eq:rel_pos_meas} can be written
\begin{align}
        \mbf{y}_k^\textrm{rel} &= 
        \mbf{C}_{ab_k} \mbf{r}^{z_2 z_1}_b + \mbf{n}_k^\textrm{rel} , \label{eq:rel_pos_meas2}
\end{align}
where $\mbf{r}^{z_2 z_1}_b$ is the known distance, the baseline, between the two receivers in $\mc{F}_b$. Some sensor suites, such as a moving baseline real-time kinematic (RTK) GPS system,  will provide both \eqref{eq:pos_meas} and \eqref{eq:rel_pos_meas2} directly \cite{Schmalzried2017}. 

% However, such sensor suites are often more costly. 

% High-end sensors, such as those that are capable of moving-base differential-GPS applications, relative position measurements can be directly obtained from the sensors.

% In Section~\ref{sec:iekfformulation},~\eqref{eq:pos_meas} is transformed to a single position measurement with a relative position measurement of the two receivers. 
%

\subsection{IEKF Formulation}
\label{sec:iekfformulation}

% The process model is formulated in the group of double direct isometries, $SE_2(3)$, with \cite{Barrau2015}
%
The kinematic equations \eqref{eq:kin_C}, \eqref{eq:kin_v}, \eqref{eq:kin_r} can collectively be written using the group of double direct isometries, $SE_2(3)$  \cite{Barrau2017TAC,Barrau2015}, to form the IEKF process model. In particular,
\begin{align}
    \mbfdot{X} = 
    \mbf{F}(\mbf{X}, \mbf{u}_b) + \mbf{X} \mbf{W}_b
\end{align}
where $\mbf{u}_b = [\mbf{u}_b^{1 ^\trans} \; \; \; \mbf{u}_b^{2 ^\trans} \; \; \; \mbf{0}]^\trans$, $\mbf{W}_b^\vee = \mbf{w}_b = [\mbf{w}_b^{1 ^\trans} \; \; \; \mbf{w}_b^{2 ^\trans} \; \; \; \mbf{0}]^\trans$,
\begin{align}
    \mbf{X} 
    & = 
    \bma{ccc}
			\mbf{C}_{ab} & \mbf{v}_a^{z_0w/a} & \mbf{r}_a^{z_0w} \\
			\mbf{0} & 1  & 0\\
			\mbf{0} & 0 & 1
	\ema \in SE_2(3) , \\
	\mbf{W}_b = \mbf{w}_b^\wedge
	& =
    \bma{ccc}
			\mbf{w}_b^{1 ^\times} & \mbf{w}_b^2 & \mbf{0} \\
			\mbf{0} & 0  & 0\\
			\mbf{0} & 0 & 0
	\ema \in \mathfrak{se}_2(3) , \\
	\mbf{F}(\mbf{X}, \mbf{u}_b) 
	& =
    \bma{ccc}
			\mbf{C}_{ab} \mbf{u}_b^{1 ^\times} & \mbf{C}_{ab} \mbf{u}_b^2 + \mbf{g}_a & \mbf{v}_a^{z_0w/a} \\
			\mbf{0} & 1  & 0\\
			\mbf{0} & 0 & 1
	\ema . \label{eq:F(X,u)}
\end{align}
The function \eqref{eq:F(X,u)} is group affine \cite{Barrau2017TAC}.
The subblocks composing the left-invariant error \eqref{eq:left_error} between $\mbf{X} \in SE_2(3)$ and $\mbfhat{X} \in SE_2(3)$ are
\begin{align}
	\delta \mbf{C}_{ab} &= \mbf{C}_{ab}^\trans \mbfhat{C}_{ab}, \\
	\delta \mbf{v}^{z_0w/a}_{a} &=  \mbf{C}_{ab}^\trans (\mbfhat{v}^{z_0w/a}_{a}-\mbf{v}^{z_0w/a}_{a}), \\
	\delta \mbf{r}^{z_0w}_{a} &= \mbf{C}_{ab}^\trans (\mbfhat{r}^{z_0w}_{a}-\mbf{r}^{z_0w}_{a}).
\end{align}
These error definitions are used to linearize the process model, resulting in a linearized model of the form (\ref{eq:proc_model}) where
\begin{align}
    \delta \mbs{\xi} =
    \bma{ccc}
        \delta \mbs{\xi}^{\theta ^\trans} & \delta \mbs{\xi}^{v ^\trans} & \delta \mbs{\xi}^{r ^\trans}
    \ema^\trans ,
\end{align}
and the Jacobians $\mbf{A}_\mathrm{c}$ and $\mbf{L}_\mathrm{c}$ are \cite{Barrau2015}
% $\mbf{A}_\mathrm{c}$ and $\mbf{L}_\mathrm{c}$ are then given by \cite{Barrau2015}
%
\begin{align}
    \mbf{A}_\mathrm{c} &=
		\bma{ccc}
			-{\mbf{u}_b^1}^\times & \mbf{0} & \mbf{0} \\
			-{\mbf{u}_b^2}^\times & -{\mbf{u}_b^1}^\times & \mbf{0} \\
			\mbf{0} & \mbf{1} & -{\mbf{u}_b^1}^\times \\
		\ema, \label{eq:Ac_SE_2(3)} \\
		\mbf{L}_\mathrm{c} &=
		\bma{ccc}
			-\mbf{1} & \mbf{0} & \mbf{0} \\
			\mbf{0} & -\mbf{1} & \mbf{0} \\
			\mbf{0} & \mbf{0} & \mbf{0} \\
		\ema . \label{eq:Lc_SE_2(3)}
\end{align}
%
% where the subscript c denotes a continuous time Jacobian. %
% The process model can be shown to be group affine. 
The Jacobian $\mbf{A}_\mathrm{c}$ in \eqref{eq:Ac_SE_2(3)} is independent of the state estimate. This is an attribute of group affine systems, a key ingredient in the IEKF framework. Additionally, $\mbf{L}_\mathrm{c}$ in \eqref{eq:Lc_SE_2(3)} happens to be constant and independent of the state estimate as well, although the IEKF framework does not guarantee this. The state-estimate independence of $\mbf{A}_\mathrm{c}$ and  $\mbf{L}_\mathrm{c}$, that does not hold for the MEKF presented in Appendix\ref{sec:MEKF}, is an advantage of the IEKF that often results in better filter performance, in particular in the transient phase \cite{Barrau2017TAC}. The continuous-time linearized process model is discretized using the method described in \cite{VanLoan1978,Farrell2008}.

% , that does not hold for the MEKF presented in Appendix~\ref{sec:MEKF}. 

% The exteroceptive measurements available are position measurements from two UWB receivers placed on the rigid body. Each

% The measurements used in the correction step are the one of the receivers and the relative position of the two receivers.
% The measurement model is defined as one position receiver and the relative position of the two position receivers. The measurement model of one position receiver is given by \eqref{eq:pos_meas}, where $\mbf{y}_k^{\textrm{pos},1}$ is

% The relative position measurement is of the form

The measurement model is composed of $\mbf{y}_k^{\textrm{pos},1}$, as defined in \eqref{eq:pos_meas}, and the relative position of two position receivers, as given in \eqref{eq:rel_pos_meas}.
 Using \eqref{eq:pos_meas} and \eqref{eq:rel_pos_meas}, the measurement model is 
\begin{align}
    \mbf{y}_k =
    \bma{c}
        \mbf{y}_k^1 \\
        \mbf{y}_k^2
    \ema , \label{eq:IEKF_meas_model}
\end{align}
where
\begin{align}
    \mbf{y}_k^1 & =
    \bma{c}
        \mbf{y}_k^{\textrm{pos},1} \\
        0 \\
        1
    \ema =
    \mbf{X}_k 
        \bma{c}
            \mbf{r}_b^{z_1z_0}\\
            0 \\
            1
        \ema + 
        \bma{c}
            \mbf{n}_k^{\textrm{pos},1} \\
            0 \\
            0
        \ema , \label{eq:IEKF_meas_model_pos} \\
    \mbf{y}_k^2 & =
    \bma{c}
        \mbf{y}_k^\textrm{rel} \\
        0 \\
        0
    \ema =
        \mbf{X}_k 
        \bma{c}
            \mbf{r}_b^{z_2z_1}\\
            0 \\
            0
        \ema + 
        \bma{c}
            \mbf{n}_k^\textrm{rel}\\
            0 \\
            0
        \ema
    . \label{eq:IEKF_meas_model_rel_pos}
\end{align}
%
% \begin{align}
%     % \mbf{y}_k =
%     \bma{c}
%         \mbf{y}_k^{\textrm{pos},1} \\
%         0 \\
%         1 \\
%         \mbf{y}_k^\textrm{rel} \\
%         0 \\
%         0 \\
%     \ema &=
%     \bma{c}
%         \mbf{X}_k 
%         \bma{c}
%             \mbf{r}_b^{z_1z_0}\\
%             0 \\
%             1
%         \ema + 
%         \bma{c}
%             \mbf{n}_k^1\\
%             0 \\
%             0
%         \ema \\[1.5em]
%         \mbf{X}_k 
%         \bma{c}
%             \mbf{r}_b^{z_2z_1}\\
%             0 \\
%             0
%         \ema + 
%         \bma{c}
%             \mbf{n}_k^2 - \mbf{n}_k^1\\
%             0 \\
%             0
%         \ema
%     \ema . \label{eq:IEKF_meas_model}
% \end{align}
%
% where $\mbf{n}_k^1 \sim \mathcal{N}(\mbf{0},\mbf{R}^{z_1w})$, and $\mbf{n}_k^2 \sim \mathcal{N}(\mbf{0},\mbf{R}^{z_2w})$. 
Both \eqref{eq:IEKF_meas_model_pos} and \eqref{eq:IEKF_meas_model_rel_pos} composing \eqref{eq:IEKF_meas_model} are left-invariant measurement models of form given in (\ref{eq:left_inv_meas}). The left-invariant innovation is then % of the form given in \eqref{eq:left_inn} is
\begin{align}
    \mbf{z}_k
    & = 
    \bma{c}
        \mbf{z}^1_k \\
        \mbf{z}^2_k
    \ema
    =
    \bma{cc}
        \check{\mbf{X}}_k^{-1} & \mbf{0} \\
        \mbf{0} & \check{\mbf{X}}_k^{-1}
    \ema
        \bma{c}
        \mbf{y}^1_k - \mbfhat{y}^1_k \\
        \mbf{y}^2_k - \mbfhat{y}^2_k
    \ema .
    \label{eq:left_inn_specific}
\end{align}
%
 % In order to linearize, (\ref{eq:left_inn}) can be expanded as 
% Linearizing \eqref{eq:left_inn} using \eqref{eq:IEKF_meas_model} yields 
The linearization of both $\mbf{z}^i_k$, $i = 1, 2$ in \eqref{eq:left_inn_specific} is accomplished using the definition of the left-invariant error \eqref{eq:left_error} and the approximation $\delta \mbf{X}_k = \exp(\delta \mbs{\xi}_k^\wedge ) \approx \mbf{1} + \delta \mbs{\xi}_k^\wedge$, as detailed in \eqref{eq:lin_inn}. The Jacobians in \eqref{eq:lin_inn} specific to \eqref{eq:left_inn_specific} are
\begin{align}
    \mbf{H}_k &=
    \bma{ccc}
		(\mbf{r}_{b}^{z_1z_0})^\times & \mbf{0} & -\mbf{1}\\
		(\mbf{r}_{b}^{z_2z_1})^\times & \mbf{0} & \mbf{0}\\
	\ema, \; \; \;
	\mbf{M}_k =
	\bma{cc}
	    \check{\mbf{C}}_{{ab}_k}^{-1} & \mbf{0} \\
	    \mbf{0} & \check{\mbf{C}}_{{ab}_k}^{-1}
	\ema,
\end{align}
where redundant rows of zeros have been removed on both the left and right-hand sides of \eqref{eq:lin_inn}.
%
% are obtained. 
The Jacobian $\mbf{H}_k$ is independent of the state-estimate, as is guaranteed by the IEKF formulation, while  Jacobian $\mbf{M}_k$ is not state-estimate independent.

\section{Simulations and Experiments} \label{sec:results}
The two-receiver IEKF is compared to a two-receiver multiplicative extended Kalman filter (MEKF), the performance of which is used as a baseline. Both the IEKF and MEKF use 9 states to represent the system. The single-receiver IEKF performance is also shown since the magnitude of the single-receiver errors help give context to the performance of the solutions of the two-receiver problem. The trajectories of the simulations and experiments are dynamic and excite sensors in all directions.
% \cite{Hong2005}.
 
\subsection{Simulations}
\begin{figure}%[b]
      \centering
      %\makebox{\includegraphics[width = 1.05\columnwidth]{RMSE_Sim.eps}}
      \includegraphics[width = 0.85\columnwidth]{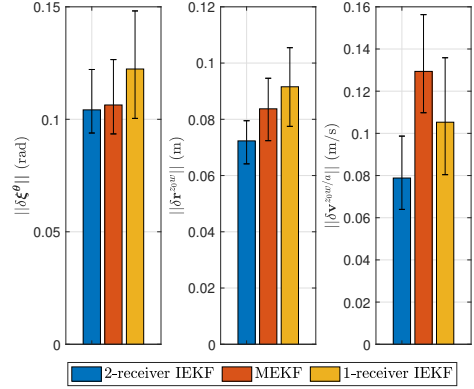}
      \vspace{-10pt}
      \caption{Attitude, position, and velocity RMSE for two-receiver IEKF and MEKF, as well as one-receiver IEKF, for 100 simulated Monte Carlo runs.\vspace{-1cm}}
      \label{fig:RMSE_sim}
\end{figure}

% The proposed approach was validated in simulation, where 
100 Monte Carlo trials were run on the same 50 second trajectory.
Each Monte Carlo trial introduces sensor noise, randomly sampled with the covariances from $\mbf{Q}_\mathrm{d} = \mathrm{diag} \Big( \mbf{Q}_\mathrm{d}^\theta, \mbf{Q}_\mathrm{d}^a  \Big)$, with $\mbf{Q}_\mathrm{d}^\theta = 0.0012^2 \times \mbf{1} \; (\mathrm{rad})^2$ and $\mbf{Q}_\mathrm{d}^a = 0.0025^2 \times \mbf{1} \; (\mathrm{m/s^2})^2$. The initial positions and velocities are randomly sampled from $\mbfhat{P}_0 = \mathrm{blkdiag} \Big( (\pi/3)^2 \times \mbf{1} \; (\mathrm{rad})^2, 0.1^2 \times \mbf{1} \; (\mathrm{m/s})^2, 0.1^2 \times \mbf{1} \; (\mathrm{m})^2 \Big)$. In order to highlight the advantage of the IEKF in the transient phase, a constant initial error of $\pi/3$ (rad) is introduced into each of the components of the attitude. The prediction step is performed at a frequency of 250 Hz, while the correction step is executed at 15 Hz. The RMSE in attitude, position and velocity for each Monte Carlo run was computed and reported in Fig.~\ref{fig:RMSE_sim}. The bars represent the average RMSE, while the upper and lower error bounds represent the 2.5 and 97.5 percentile, respectively, meaning that 95\% of trials lie between these bounds. 

Fig.~\ref{fig:RMSE_sim} shows that the two-receiver IEKF outperforms the two-receiver MEKF and single-receiver IEKF in its estimates of all states. These results are consistent with the literature, since the advantage of state-independent Jacobians emanates when the state estimates are highly erroneous, as is the case when the initial error is large. Thus, the steady state performance of the two-receiver IEKF and two-receiver MEKF should be similar, while the transient performance of the two-receiver IEKF should show improvement over the two-receiver MEKF \cite{Barrau2017}. Fig.~\ref{fig:RMSE_sim} shows that the improvement in attitude is small, while the greater advantage arises in position and velocity estimates. The advantage arising from the two-receiver IEKF framework appears in the comparison of RMSE in velocity, where the single-receiver IEKF has lower RMSE than the two-receiver MEKF.

\begin{figure}[b]
      \centering
      \vspace{-20pt}
      %\makebox{\includegraphics[width = 1.05\columnwidth]{NIS_Comp.eps}}
      \includegraphics[width = 0.85\columnwidth]{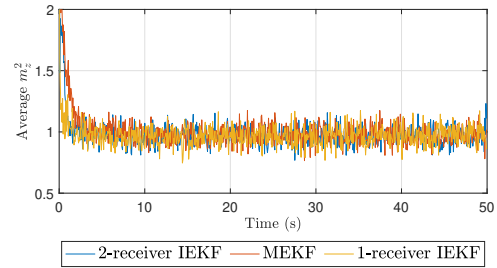}
      \vspace{-22pt}
      \caption{Comparison of simulation normalized average NIS for two-receiver IEKF, MEKF and one-receiver IEKF for 100 Monte Carlo runs.}%\vspace{-0.5cm}}
      \label{fig:NIS_Comp}
\end{figure} 

The former results do not give any indication of the consistency of the filters. For this, the average NIS test was used. The comparison is presented in Fig.~\ref{fig:NIS_Comp}. All filters were found to become consistent within less than 5 seconds, with the initial inconsistency caused by the error introduced in the attitude initialization.

\vspace{-4pt}
\subsection{Experiment}
\label{sec:experiments}
\begin{figure}%[b]
      \centering
      %\makebox{\includegraphics[width = 1.05\columnwidth]{RAL_Pavlasek_Source/Exp_setup2.jpg}}
      \includegraphics[width = 0.8\columnwidth]{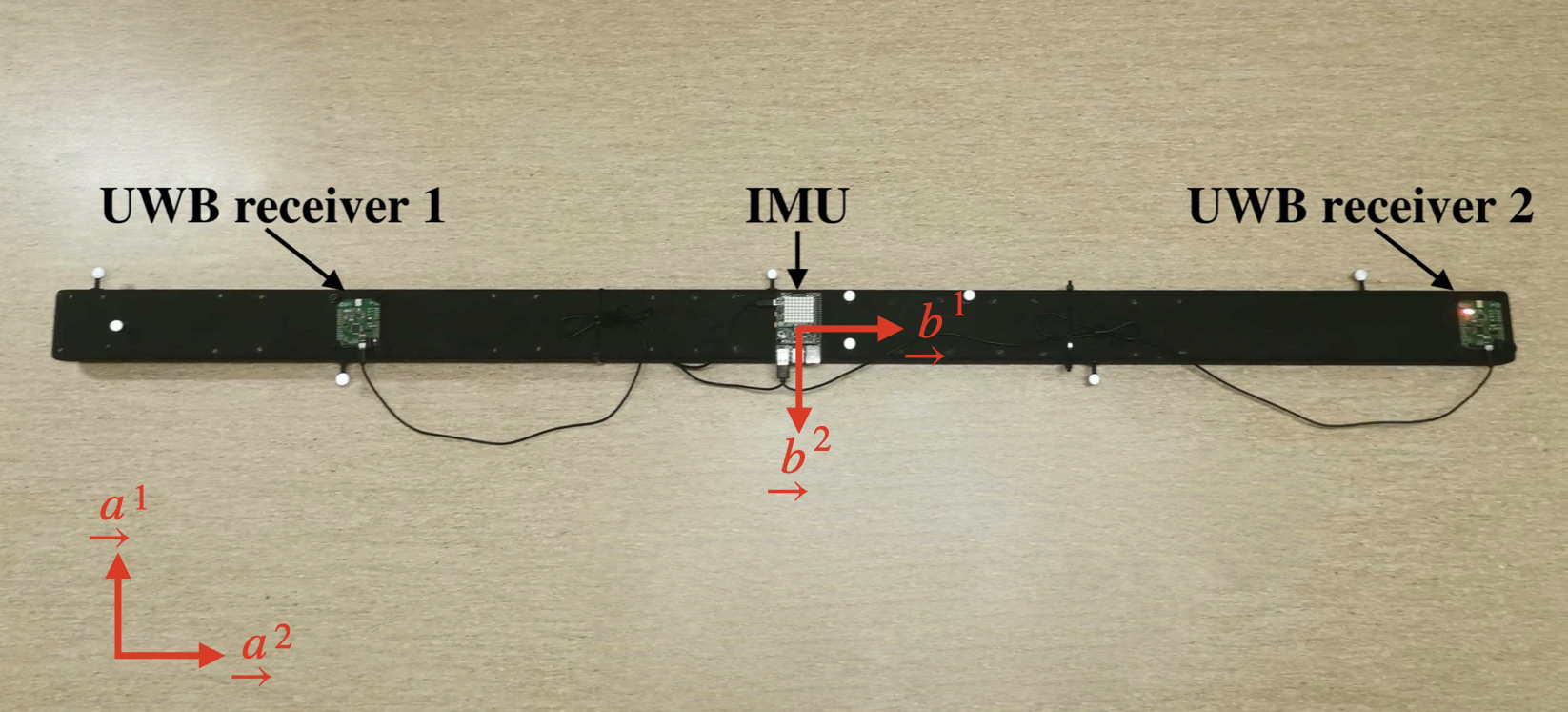}
      \vspace{-5pt}
      \caption{Experimental setup. Two UWB position receivers and one IMU mounted on a rigid body that is moved by hand to collect data. \vspace{-1cm} }
      \label{fig:exp_setup}
\end{figure}

The proposed approach was further validated through an experiment. An IMU and two UWB receivers were mounted on a rigid body, shown in Fig.~\ref{fig:exp_setup}, with the two UWB receivers placed 1.80 m apart. Five UWB anchors were placed at various heights around a room where the experiments were conducted. Trials of approximately two minutes were recorded, during which the rigid body was moved around a volume measuring 5~m $\times$ 4~m $\times$ 2~m. A motion capture system was used to record ground truth measurements. The accelerometer and gyroscope measurements were collected at a frequency of 250~Hz, while the UWB position measurements arrived at a frequency of about 17~Hz.

% 4~m $\times$ 2~m $\times$ 2~m.

\begin{figure}%[t]
      \centering
      \makebox{\includegraphics[width = 0.9\columnwidth]{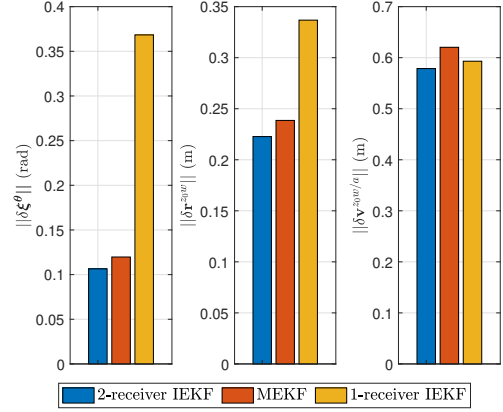}}
      \vspace{-10pt}
      \caption{Comparison of experiment RMSE in attitude, position, and velocity for two-receiver IEKF, MEKF, and one-receiver IEKF.\vspace{-0.5cm} }
      \label{fig:RMSE_exp}
\end{figure}

The process noise power spectral density (PSD) matrices were computed using the Allan variance method, generated using static IMU data, according to \cite{Woodman2007}. The measurement noise covariance matrix was computed through comparison of the recorded measurements and motion capture ground truth measurements. The resulting PSD and covariance matrices are
\begin{align}
    \mbs{Q}(t) &= \mathrm{diag}\Big( \mbs{Q}^\omega (t), \mbs{Q}^\mathrm{a} (t) \Big), \\
    \mbs{Q}^\omega (t) &= \mathrm{diag}\Big( 2.0^2, 2.0^2, 1.8^2 \Big) \times 10^{-4} \; 
    (\mathrm{rad})^2,\\
    \mbs{Q}^\mathrm{a} (t) &= \mathrm{diag}\Big( 1.7^2, 1.5^2, 2.4^2 \Big) \times 10^{-2} \; (\mathrm{m/s^2})^2,\\
    \mbf{R} &= \mathrm{diag}\Big( \mbf{R}^{z_1w}, \mbf{R}^{z_2w} \Big), \\
    \mbf{R}^{z_1w} &= \mathrm{diag}\Big( 1.3^2, 1.1^2, 1.9^2 \Big) \times 10^{-2} \; (\mathrm{m})^2,\\
    \mbf{R}^{z_2w} &= \mathrm{diag}\Big( 1.9^2, 1.6^2, 2.6^2 \Big) \times 10^{-2} \; (\mathrm{m})^2.
\end{align}
As in simulation, the attitude was initialized with an error of $\pi/3$ (rad) to demonstrate the advantage of the IEKF in situations in which there is large uncertainty on the initial attitude. The RMSE for each of the filters is presented in Fig.~\ref{fig:RMSE_exp}.  It should be noted that ground truth velocity was not measured directly, but rather computed using the ground truth positions. This process introduces some error into the ground truth velocity, which is used to compute the error in the velocity estimates.

% Unlike in simulation, velocity ground truth is not available for the experiment and thus the RMSE in velocity is not presented.

The filter performance on experiment data emulates the simulation performance. The two-receiver IEKF attitude estimate improves greatly over the single-receiver IEKF, but show only slight improvement over the two-receiver MEKF. This improvement arises in the transient phase where the effect of having state-dependent Jacobians impacts the MEKF but not the IEKF. The two-receiver IEKF and MEKF perform equivalently at steady state, which is consistent with previous literature. It should be noted that a transient phase does not only occur during initialization, but can also occur if a sensor is unavailable (that is, ``drops out") for an extended period of time.

\subsection{Receiver Spacing Study}

\begin{figure}%[t]
      \centering
      \makebox{\includegraphics[width = 1\columnwidth]{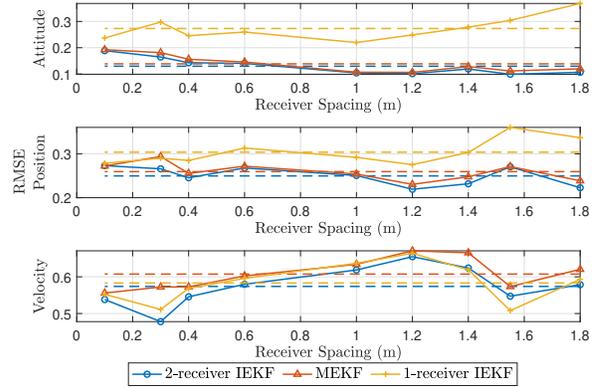}}
      \vspace{-5pt}
      \caption{RMSE of each filter and the two-receiver IEKF in attitude, position, and velocity for various tag spacings.\vspace{-0.5cm}
      %Percent difference between RMSE of each filter and the two-receiver IEKF in attitude and position for various tag spacings.\vspace{-0.5cm}
      }
      \label{fig:RMSE_comp}
\end{figure}

\begin{figure}%[t]
      \centering
      \makebox{\includegraphics[width = 0.95\columnwidth]{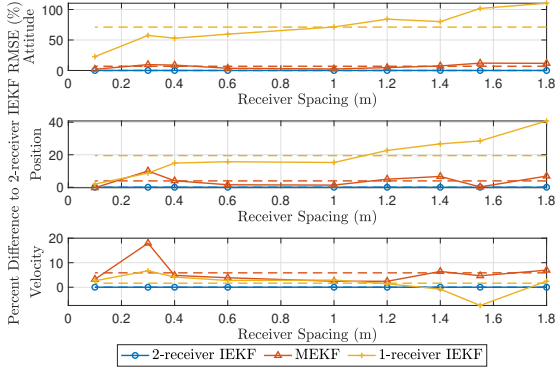}}
      \vspace{-5pt}
      \caption{Percent difference between RMSE of each filter and the two-receiver IEKF in attitude, position, and velocity for various tag spacings.\vspace{-0.5cm}
      }
      \label{fig:RMSE_perc_diff}
\end{figure}

% After validating the filter in simulation and an experiment, 
The distance between receivers was studied experimentally using the platform shown in Fig.~\ref{fig:exp_setup}. % The same type of experiment was performed, with the receivers placed at various locations along the rigid body.
The receiver placement was decreased between experimental runs. 
For each of the experiment runs, the three filters were compared using RMSE as a performance metric. 
%The resulting percent differences between each filter and the two-receiver IEKF are shown in Fig.~\ref{fig:RMSE_comp}.
The resulting RMSEs of each filter are shown in Fig.~\ref{fig:RMSE_comp}.
While reporting the single-receiver IEKF performance as a function of receiver spacing may seem imprecise, the receiver spacing is used to categorize the various experiment runs. While the RMSE of the single-receiver should be consistent across runs, certain inconsistencies arise from factors such as the manual trajectory generation and hardware flaws during data collection. Thus, the single-receiver RMSE are computed and reported for each run. The same data is presented in terms of percent differences between each filter and the two-receiver IEKF in Fig.~\ref{fig:RMSE_perc_diff}.

The two-receiver IEKF has the greatest advantage over the single-receiver IEKF in attitude estimates, in which a maximum improvement of 110\%, and a minimum improvement of 23\% was observed. With greater receiver spacing, the advantage introduced by the addition of a second receiver becomes increasingly evident in the position and attitude estimates. Even with the receivers spaced 0.1 m apart, the addition of a second receiver enhances the attitude estimates by 23\%, and the position estimates by 2\%. Over all spacings, the two-receiver IEKF performance improved attitude estimates by 71\% over the single-receiver IEKF and by 7\% over the two-receiver MEKF. The position estimates were improved by 19\% over the single-receiver IEKF and 4\% over the two-receiver MEKF.

\subsection{Extension to the Unscented Kalman Filtering Framework}

% Using this formulation, the measurement model presented in the present work can be used with a UKF. 

\begin{figure}%[b]
      \centering
      %\makebox{\includegraphics[width = 1.05\columnwidth]{NIS_Comp.eps}}
      \includegraphics[width = 1\columnwidth]{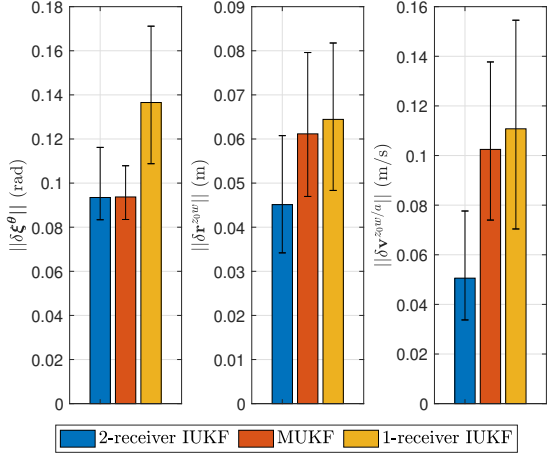}
      \vspace{-10pt}
      \caption{Attitude, position, and velocity RMSE for two-receiver IUKF and MUKF, as well as one-receiver IUKF, for 100 simulated Monte Carlo runs.\vspace{-0.5cm}}
      \label{fig:UKF_sim}
\end{figure}

While the focus of this work is on the use of the IEKF, the invariant measurement, invariant error, invariant innovation, and group-affine process model can be used within an invariant UKF (IUKF) framework  \cite{Brossard2017IUKF}. Simulation results comparing the performance of the two-receiver IUKF, the two-receiver multiplicative UKF (MUKF), and the single-receiver IUKF are presented in Fig.~\ref{fig:UKF_sim}, without a derivation of the filter equations. The derivation of the filters equations follows that of \cite{Brossard2017IUKF}.

The average RMSE in attitude of the IUKF is slightly lower than that of the UKF, but the IUKF has slightly larger upper and lower percentile bounds than the UKF. The IUKF definitively outperforms the MUKF in terms of position and velocity RMSE. Also, given the same parameters, the three UKF variants, with results presented in Fig.~\ref{fig:UKF_sim}, outperform the corresponding EKF variant, with results presented in Fig.~\ref{fig:RMSE_sim}. Therefore, should adequate computing resources be available such that a UKF variant can be used, and performance beyond what the two-receiver versions of the MEKF, MUKF, and IEKF can provide is required, the IUKF is attractive. 

% in situations where enhanced performance is required beyond what the two-receiver versions of the MEKF, the MUKF, and even the IEKF can provide, the IUKF is attractive. 

% However, the IUKF is not without it's own challenges, such as being more computationally burdensome than the EKF

\section{Conclusion} \label{sec:conclusion}

This paper's novel contribution is demonstrating how to use the IEKF to estimate position, velocity, and attitude, that being the extended pose, using two position receivers and IMU data. The two-receiver problem is shown to be left invariant, and all other requirements of the invariant framework are met, enabling the use of the IEKF. The proposed two-receiver IEKF is compared in simulation and experiment to a two-receiver MEKF and a single-receiver IEKF. The two-receiver IEKF was found to have improved performance over the MEKF when the initialization error was large, as is consistent with literature. 

% In order to fit within this structure, a system must have left- or right-invariant measurement models, with corresponding left- or right-invariant error definitions and innovations, as well as group affine process models to guarantee state-independent Jacobians. The two-receiver IEKF was shown to fit within this framework. It was compared in simulation and in an experiment to an MEKF and to a single-receiver IEKF. The two-receiver IEKF was found to have improved performance over the MEKF when the initialization error was large, as is consistent with literature. 

% The primary goal of this paper is to demonstrate the advantage introduced by the addition of a second position receiver in the IEKF framework when estimating extended pose. 

% \colour{red}{Word smith this a bit: The addition of a second receiver resulted in a large advantage over the single-receiver IEKF in attitude estimates. The advantage of the two-receiver IEKF over the other filters increased as the two receivers were placed closer further apart, with the exception of velocity estimates.}

\appendices
\section*{Appendix}
\subsection{MEKF} \label{sec:MEKF}
The kinematics in \eqref{eq:kin_C}, \eqref{eq:kin_v}, \eqref{eq:kin_r} are framed in the multiplicative extended Kalman filter (MEKF) framework. Here, the linearized process model is of the form,
\begin{align} \label{eq:EKF_proc_model}
    \delta \mbfdot{x} = \mbf{A}_\mathrm{c} \delta \mbf{x} + \mbf{L}_\mathrm{c} \delta \mbf{w},
\end{align}
where % the state is,
\begin{align}
    \delta \mbf{x} =
    \bma{ccc}
        \delta \mbs{\xi}^{\theta ^\trans} & \delta \mbf{v}_a^{z_0w/a  ^\trans} & \delta \mbf{r}_a^{z_0w  ^\trans}
    \ema^\trans,
\end{align}
and $\exp(\delta \mbs{\xi}^{\theta ^\times}) = \mbf{C}_{ab} \mbfhat{C}_{ab}^\trans$, $\delta \mbf{v}_a^{z_0w/a} = \mbf{v}_a^{z_0w/a} - \mbfhat{v}_a^{z_0w/a}$, and $\delta \mbf{r}_a^{z_0w} = \mbf{r}_a^{z_0w} - \mbfhat{r}_a^{z_0w}$.
The process model Jacobians are then
\begin{align}
    \mbf{A}_\mathrm{c} &= 
    \bma{ccc}
        -{\mbf{u}_b^1}^\times & \mbf{0} & \mbf{0} \\
        -\mbfhat{C}_{ab}{\mbf{u}_b^2}^\times & \mbf{0} & \mbf{0} \\
        \mbf{0} & \mbf{1} & \mbf{0}
    \ema, \; \; \;
    \mbf{L}_\mathrm{c} =
    \bma{ccc}
        -\mbf{1} & \mbf{0} & \mbf{0} \\
        \mbf{0} & -\mbfhat{C}_{ab} & \mbf{0} \\
        \mbf{0} & \mbf{0} & \mbf{0}
    \ema.
\end{align}
Note that both $\mbf{A}_\mathrm{c}$ and $\mbf{L}_\mathrm{c}$ are state dependent.

The MEKF measurement models are identical to those used in the IEKF, with
\begin{align}
    \hspace{-10pt}
    \bma{c}
        \mbf{y}^1_k\\
        \mbf{y}^2_k
    \ema =
    \bma{c}
        \mbf{y}_k^{\textrm{pos},1}\\
        \mbf{y}_k^{\textrm{rel}}
    \ema =
    \bma{c}
        \mbf{r}^{z_1w}_{a_k} + \mbf{n}_k^1\\
        \mbf{r}^{z_2w}_{a_k} - \mbf{r}^{z_1w}_{a_k} + \mbf{n}_k^1 + \mbf{n}_k^2
    \ema.
\end{align}
The MEKF innovation is then
\begin{align}
    \delta \mbf{y} &=
    \bma{c}
        \mbf{r}_{a_k}^{z_1w} + \mbf{n}^1_k - \check{\mbf{C}}_{ab}\mbf{r}_b^{z_1z_0} - \check{\mbf{r}}_a^{z_0w}\\
        \mbf{r}_{a_k}^{z_2w} - \mbf{r}_{a_k}^{z_1w} + \mbf{n}^1_k + \mbf{n}^2_k - \check{\mbf{C}}_{ab}\mbf{r}_b^{z_2z_1}
    \ema,
\end{align}
which can be linearzied and written in the form
\begin{align} \label{eq:EKF_meas_model}
    \delta \mbf{y}_k = \mbf{H}_k\delta \mbf{x}_k + \mbf{M}_k\delta \mbf{n}_k,
\end{align}
where
\begin{align}
    \mbf{H}_k &=
    \bma{ccc}
        \check{\mbf{C}}_{{ab}_k} {\mbf{r}_b^{z_1z_0}}^\times & \mbf{0} & \mbf{1} \\
        \check{\mbf{C}}_{{ab}_k} {\mbf{r}_b^{z_2z_1}}^\times & \mbf{0} & \mbf{0}
    \ema, \; \; \;
    \mbf{M}_k =
    \bma{cc}
        \mbf{1} & \mbf{0} \\
        \mbf{0} & \mbf{1}
    \ema .
\end{align}
\section*{Acknowledgment}
The authors are most grateful to Charles Cossette and Mohammed Shalaby for their help collecting the experimental datasets used in this paper.

%%%%%%%%%%%%%%%%%%%%%%%%%%%%%%%%%%%%%%%%%%%%%%%%%%%%%%%%%%%%%%%%%%%%%%%%%%%%%%%%
%\bibliographystyle{IEEEtran}
%\bibliography{IEEEabrv,IEEEexample}
{\AtNextBibliography{\small}
\balance
\printbibliography}
\end{document}